\documentclass[11pt,a4paper]{article}
\usepackage{times,latexsym}
\usepackage{url}
\usepackage[T1]{fontenc}
\usepackage{times}
\usepackage{amsmath}
\usepackage{multirow}
\usepackage{latexsym}
\usepackage{color}
\usepackage{makecell}
\usepackage{adjustbox}
\usepackage{amssymb}
\usepackage{parskip}
\usepackage{algorithm}
\usepackage{algpseudocode}

\usepackage[acceptedWithA]{tacl2021v1}

\usepackage{xspace,mfirstuc,tabulary}

\usepackage{float}
\usepackage[utf8]{inputenc}

\newif\iftaclinstructions
\taclinstructionsfalse 
\iftaclinstructions
\renewcommand{\confidential}{}
\renewcommand{\anonsubtext}{(No author info supplied here, for consistency with
TACL-submission anonymization requirements)}
\newcommand{\instr}

\fi

\iftaclpubformat 

\else

\fi

\title{Rank-Aware Negative Training for Semi-Supervised Text Classification}

 \author{
   Ahmed Murtadha \Thanks{Corresponding author}$^\diamond$ 
   Shengfeng Pan$^\diamond$
   Wen Bo$^\diamond$
   Jianlin Su$^\diamond$
   Xinxin Cao$^\dagger$\\
   \textbf{Wenze Zhang}$^\diamond$
   \textbf{Yunfeng Liu}$^\diamond$
   \\
   \ \\
   $^\diamond$ Zhuiyi AI Lab, Shenzhen, Guangdong, China
   \\
   \texttt{\{a.murtadha,nickpan,brucewen,bojonesu,wen,glenliu\}@wezhuiyi.com}
   \\ 
   $^\dagger$Northwestern Polytechnical University, Xi'an, Shaanxi, China\\
   \texttt{caoxinxin@mail.nwpu.edu.cn}
 }

\date{}

\begin{document}
\maketitle
\begin{abstract}
  Semi-supervised text classification-based paradigms (SSTC) typically employ the spirit of self-training. The key idea is to train a deep classifier on limited labeled texts and then iteratively predict the unlabeled texts as their pseudo-labels for further training. However, the performance is largely affected by the accuracy of pseudo-labels, which may not be significant in real-world scenarios. 
  This paper presents a Rank-aware Negative Training (RNT) framework  to address SSTC in learning with noisy label manner.
  To alleviate the noisy information, we adapt a reasoning with uncertainty-based approach to rank the unlabeled texts based on the evidential support received from the labeled texts. 
  Moreover, we propose the use of negative training to train RNT based on the concept that ``the input instance does not belong to the complementary label''.  A complementary label is randomly selected from all labels except the label on-target. Intuitively, the probability of a true label serving as a complementary label is low and thus provides less noisy information during the training, resulting in better performance on the test data.
  Finally, we evaluate the proposed solution on various text classification benchmark datasets. Our extensive experiments show that it consistently overcomes the state-of-the-art alternatives in most scenarios and achieves competitive performance in the others. The code of RNT is publicly available on \href{https://github.com/amurtadha/RNT}{GitHub}.
\end{abstract}

\section{Introduction}

Text classification task aims to associate a piece of text with a corresponding class that could be a sentiment, topic, or category. 
With the rapid development of deep neural networks, text classification has experienced a considerable shift towards the pre-trained language models (PLMs)  
\cite{devlin2018bert,yang2019xlnet,liu2019roberta,lewis2019bart}. 
Overall, PLMs are first trained on massive text corpora (e.g., Wikipedia) to learn contextual representation, followed by a fine-tuning step on the downstream tasks  \cite{li2021semi,chen2022dual,tsai2022contrast,murtadha2022bert}. The improvement of these approaches heavily relies on high-quality labeled data. However, labeling data is labor-intensive and may not be readily available in real-world scenarios. 
To alleviate the burden of labeled data, Semi-Supervised Text Classification (SSTC) typically refers to leveraging unlabeled texts to perform a particular task.
SSTC-based approaches commonly attempt to exploit the consistency between instances under different perturbations  \cite{li2019dividemix}.

Earlier SSTC-based approaches adopt various data augmentation techniques via back-translation. They employ consistency loss between the predictions of unlabeled
texts and corresponding augmented texts by translating the text into a targeted language and then translating it back to the source language  \cite{miyato2018virtual,xie2020unsupervised,chen2020mixtext}. However, the performance of these approaches requires an additional Neural Machine Translation (NMT), which may not be accurate and bothersome in real-world scenarios. Recently, SSTC has experienced a shift toward self-training, and PLMs fine-tuning  \cite{li2021semi,tsai2022contrast}. 
The basic idea is to fine-tune PLMs on the labeled data and iteratively employ prediction on the unlabeled data as pseudo-labels for further training. However, the pseudo-labels are treated equally likely to the truth labels and thus may lead to the error accumulation  \cite{zhang2021understanding,arazo2020pseudo}.

In this paper, we propose a Rank-aware Negative Training (RNT) framework to address SSTC under learning with noisy labels settings. To alleviate the domination of noisy information during training, we adopt reasoning with an uncertainty-based approach to rank the unlabeled texts by measuring  their shared features, also known as evidential support, with the labeled texts.
Eventually, the shared features  that serve as a medium to convey knowledge from labeled texts (i.e., evidence) to the unlabeled texts (i.e., inference) are regarded as belief functions to reason about the degree of noisiness. These belief functions are combined to reach a final belief about the text being mislabeled. In other words, we attempt to discard the texts whose pseudo-labels may introduce inaccurate information to the training process.    

Moreover, we propose using negative training (NT)  \cite{kim2019nlnl} to robustly train with potential noisy pseudo-labels. Unlike positive training, NT is an indirect learning method that trains the network based on the concept that ``the input sentence does not belong to the complementary label'', whereas a complementary label is randomly generated from the label space except the label of the sentence on-target. Considering the AG News dataset, given a sentence annotated as \textit{sport}, the complementary label is randomly selected from all labels except \textit{sport} (e.g., \textit{business}).
Intuitively, the probability of a true label serving as a complementary label is low and thus can reduce the noisy information during the training process. Finally, we conduct extensive experiments on various text classification benchmark datasets with different ratios of labeled examples, resulting in better performance on the test data. Experimental results suggest that RNT can mostly outperform the SSTC-based alternatives. Moreover, {  it has been empirically }shown that RNT can perform better than PLMs fine-tuned on sufficient labeled examples. 

In brief, the main contributions are three-fold: 
\begin{itemize}
	\item We propose a rank-aware negative training framework, namely RNT, to address the semi-supervised text classification problem as learning with the noisy labels manner. 
	\item We introduce reasoning with an uncertainty-based solution to discard texts with the potential noisy pseudo-labels by measuring evidential support received from the labeled texts.
	\item We evaluate the proposed solution on various text classification benchmark datasets. Our extensive experiments show that it consistently overcomes the state-of-the-art alternatives in most cases and achieves competitive performance in others.
\end{itemize}

\section{Related Work}\label{sec:related_work}
This section reviews the existing solutions of SSTC task and learning with noisy labels.

\textbf{Text Classification.}
Text classification  aims at assigning  a given document to a number of semantic categories, which could be a sentiment, topic or aspect  \cite{hu2004mining,liu2012sentiment,Schouten2016tkde}. 
Earlier solutions were usually equipped with a deep memory or  an attention mechanism to learn semantic representation in response to a given category  
\cite{socher2013recursive,zhang2015character,wang2016attention,ma2017interactive,chen2017recurrent,johnson2017deep,conneau2017very,song2019attentional,ahmed2020constructing,tsai2022contrast}.
Recently, many NLP tasks have experienced a considerable shift towards fine-tuning the pre-trained language models (PLMs)   
\cite{devlin2018bert,yang2019xlnet,liu2019roberta,zaheer2020big,chen2022dual,tsai2022contrast,murtadha2022bert}. 
Despite the effectiveness of these approaches, the performance heavily relies on the quality of the labeled data, which requires intensive human labor.

\textbf{Semi-supervised text classification.}
Partially supervised text classification, also known as learning from Positive and Unlabeled (PU) examples, aims at building a classifier using P and U in the absence of negative examples to classify the Unlabeled examples \cite{liu2002partially,li-etal-2010-negative,liu2011partially}. Recent SSTC approaches primarily focus on exploiting the consistency in the predictions for the same samples under different perturbations.
 \citet{miyato2016adversarial} established virtual adversarial training that perturb word embeddings to encourage consistency between perturbed embeddings. Variational auto-encoders-based approaches  \cite{yang2017improved,chen2018variational,gururangan2019variational} attempted to reconstruct instances and utilized the latent variables to classify text. Unsupervised data augmentation (UDA)  \cite{xie2020unsupervised} performed consistency training by making features consistent between back-translated instances.
However, these methods mostly require additional systems (e.g., NMT back-translation), which may be bothersome in real-world scenarios.
\citet{mukherjee2020uncertainty} and \citet{tsai2022contrast} introduced uncertainty-driven self-training-based solutions to select samples and performed self-training on the selected data. An iterative framework  \cite{ma2021sent}, named SENT, proposed to address distant relation extraction via negative training. Self-Pretraining   \cite{karisani2021semi} was introduced to employ an iterative distillation procedure to cope with the inherent problems of self-training. SSTC-based approaches and their limitations are well-described by  \citet{van2020survey} and \citet{yang2021survey}.   Recently, S$^2$TC-BDD  \cite{li2021semi} was introduced to balance the label angle variances (i.e., the angles between deep representations of texts and weight vectors of labels), also called the margin bias.
Despite the effectiveness of these methods, the unlabeled instances contribute equally likely to the labeled ones; therefore, the performance heavily relies on the quality of pseudo-labels. Unlikely, our proposed solution addresses the SSTC task as a learning under noisy label settings problem. Since the pseudo-labels are automatically labeled by the machine, we thus regard them as noisy labels and introduce a ranking approach to filter the highly risky mislabeled instances. To alleviate the noisy information resulting from the filtering process, we use negative training that performs classification based on the concept that ``the input instance does not belong to the complementary label''. 

\textbf{Learning with noisy labels.}
Learning with noisy data has been extensively studied, especially in computer vision community. The existing solutions introduced various methods to relabel the noisy samples in order to correct the loss function. To this end, several relabeling methods have been introduced to treat all samples equally to model the noisy ones, including directed graphical models  \cite{xiao2015learning}, conditional random fields  \cite{vahdat2017toward}, knowledge graph  \cite{8237473}, or deep neural networks \cite{8100179,8578669}. However, they were built based on semi-supervised learning, where access to a limited number of clean data is required.  \citet{pmlr-v80-ma18d} introduced a bootstrapping method to modify the loss with model predictions by exploiting the dimensionality of feature subspaces.   \citet{8099723}  proposed to estimate the label corruption matrix for loss correction.
Another direction of research on loss correction investigated two approaches, including reweighting training samples and separating clean and noisy samples  \cite{pmlr-v97-thulasidasan19a,pmlr-v97-konstantinov19a}.  \citet{pmlr-v97-shen19e} have claimed that the deep classifier normally learns the clean instances faster than the noisy ones. Based on this claim, they consider instances with smaller losses as clean ones. A negative training technique  \cite{kim2019nlnl} was introduced to train the model based on the complementary label, which is randomly generated from the label space except for the label on-target. The goal is to encourage the probability to follow a distribution such that the noisy instances are largely distributed in low-value areas and the clean data are generally distributed in high-value areas to facilitate the separation process.  \citet{NEURIPS2018_a19744e2} proposed to jointly train two networks that select small-loss samples within each mini-batch to train each other. Based on this paradigm, \citet{pmlr-v97-yu19b} introduced to update the network on disagreement data to keep the two networks diverged. In this paper, we leverage a robust negative loss  \cite{kim2019nlnl} for noisy data training.

\section{Ranked-aware Negative Training}\label{sec:approach}
\begin{figure}
	\centering
	\includegraphics[scale=.75]{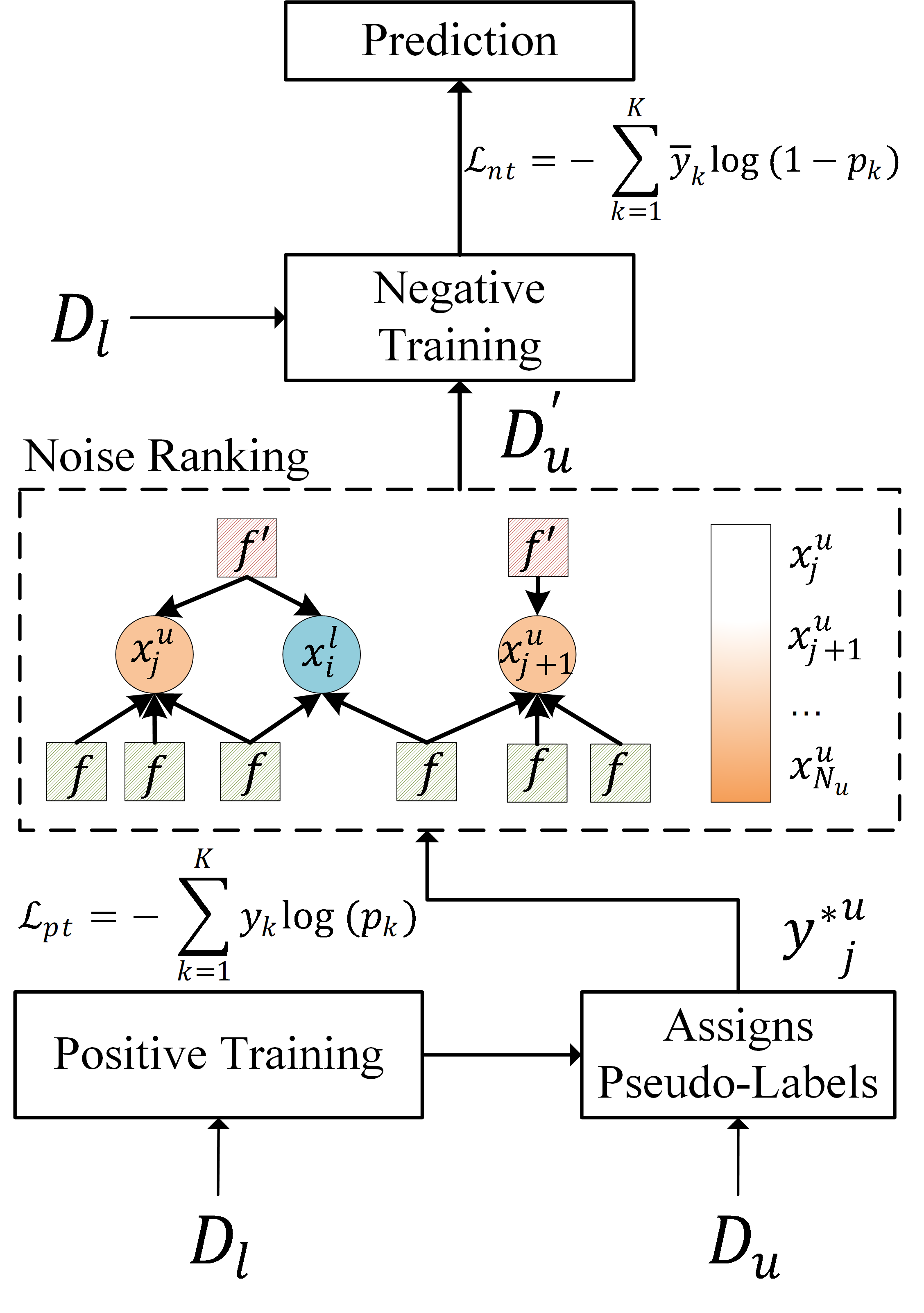}	
	\caption{An example of the proposed framework. $D_l$, $D_u$, and $D'_u$ denote labeled set, unlabeled set and filtered unlabeled set, respectively. Briefly, RNT consists of three key steps: (1) Training with PT on limited labeled texts and then iteratively predicting the unlabeled texts as their pseudo-labels; (2) Measuring the evidential support based on the learned embedding space of PT to estimate the degree of noise; (3) Training with NT on the mixture of clean and filtered data.}
	\label{fig:framework}
\end{figure}

This section describes the proposed framework, namely Rank-aware Negative Training (RNT), for semi-supervised text classification. An example of RNT is depicted in Figure \ref{fig:framework}.
Suppose we have a training dataset $D$ consisting of a limited labeled set $D_l$ and a large unlabeled set $D_u$. We follow the pseudo-labels method introduced by \citet{lee2013pseudo} to associate $D_u$ with pseudo-labels based on the concept of positive training. { Simply, we fine-tune the pre-trained language models (e.g., BERT) on the $D_l$ set. It is noteworthy that we use BERT for a fair comparison, while other models can be used similarly. As the pseudo-labels are not manually annotated, we propose ranking the texts based on their potential for mislabeling to identify and discard the most risky mislabeled texts. Specifically, we first capture the shared information (i.e., we refer to it as the evidential support) between the labeled and unlabeled instances. Then, we measure the amount of support that an unlabeled instance receives from the labeled instances being correctly labeled. We denote the filtered set as $D'_u$ in Figure\ref{fig:framework}.} Finally, we train on both $D_l$ and $D'_u$ through the concept of the negative training. Next, we describe the framework in detail.

\subsection{Task Description} 
\textbf{\textit{Semi-Supervised Text Classification (SSTC)}.}
Let $D$ be the training dataset consisting of a limited labeled set $D_l =\{(x_i^l,y_i^l )\}^{i=N_l}_{i=1}$ and a large unlabeled text set $D_u =\{(x_j^u)\}^{j=N_u}_{j=1}$, where $x_i^l$ and  $x_j^u$ denote the input sequences of labeled and unlabeled texts, respectively, and $y_i^l \in \{0,1\}^K$ represents the corresponding one-hot label vector of $x_i^l$. The goal is to learn a classifier that leverages both $D_l$ and $D_u$ to better generate in the inference step, also known as inductive SSTC.  

\subsection{Positive and Negative Training} \label{sec:pl}
\textbf{Positive Training (PT).} A typical method of training a model with a given input instance and the corresponding labels is referred to as positive training (PT). In other words,  the model is trained based on the concept that ``the input instance belongs to this label''. 
Considering a multi-class classification problem, let $x \in \mathcal{X}$ be an input,   $y\in \{0,1\}^K $ be a c-dimension
one-hot vector of its label.
The training objective $f(x;\theta)$ aims to map  the input instance  to the $k$-dimensional score space $f : \mathcal{X} \rightarrow \mathbb{R}^k$, where $\theta$ is the set of parameters. To achieve this, PT uses the cross-entropy loss function defined as follows: 
\begin{equation}\label{eq:pl}
	\mathcal{L}_{pt}= - \sum_{k=1}^{K} y_k \log(p_k),
\end{equation}
where $p_k$ denotes the probability of the $k^{th}$ label. Equation \ref{eq:pl} satisfies the claim of PT to optimize the probability value corresponding to the given
label as 1 $(p_k \rightarrow 1)$.  

\textbf{Negative Training (NT).} Unlike PT, the model is trained based on the concept that ``the input text does not belong to this label''. Specifically, given an input text $x$ with a label $y\in \{0,1\}^K $, a complementary label $\overline{y}$ is generated by randomly sampling from the label space except $y$ (e.g., $\overline{y}  \in \mathbb{R}\backslash \{y\}$). The cross-entropy loss function of NT is defined as follows.

\begin{equation}\label{eq:nt}
	\mathcal{L}_{nt}= - \sum_{k=1}^{K} \overline{y_k} \log(1-p_k).
\end{equation}
\begin{figure*}
	\centering
	\includegraphics[scale=.55]{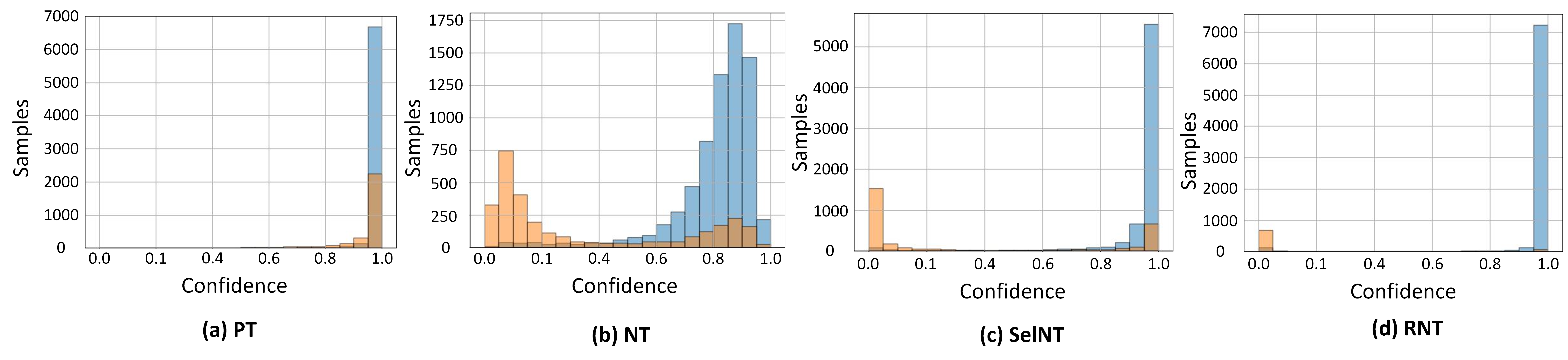}	
	\caption{{A histogram of PT, NT and RNT data training distribution conducted on AG NEWS dataset with random 30\% noisy-labels, in which blue represents the clean data and orange indicates the noisy data. SelNT further trains the model with only the samples having NT confidence over $\frac{1}{K}$, where $K$ denotes the number of classes.} }
	\label{fig:pt_nt}
\end{figure*}

\begin{figure}
	\centering
	\includegraphics[scale=.5]{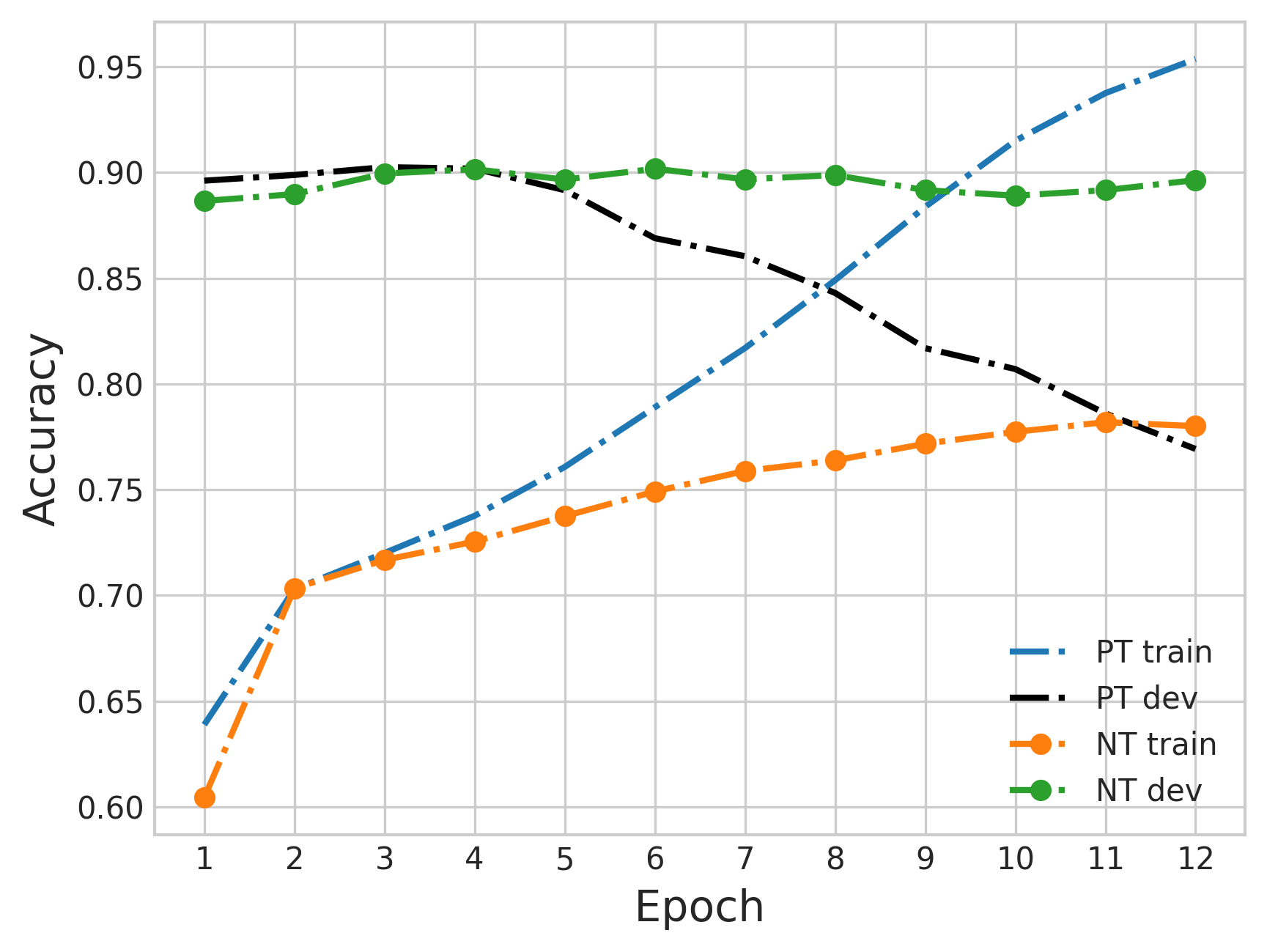}	
	\caption{A comparison between PT and NT techniques trained on AG News dataset corrupted with randomly 30\% of symmetric noise.  The accuracy of PT on the clean Dev data increases in the early stage. However, overfiting to the noisy training examples results in gradual inaccurate performance on the clean Dev data.}
	\label{fig:pt_nt_performace}
\end{figure}

To illustrate the robustness of PT and NT against noisiness, we train both techniques on AG News dataset corrupted with  randomly 30\% of symmetric noise (i.e., associating the instance with a random label).
In terms of confidence (i.e., the probability of the true class),  we  illustrate the histogram of the training data after PT and NT in Figure \ref{fig:pt_nt}. As can be seen, with PT in Figure \ref{fig:pt_nt} (a), the confidence of both clean and noisy instances increases simultaneously. With NT in Figure \ref{fig:pt_nt} (b), in contrast, the noisy instances yield much lower confidence compared to the clean ones and thus discourages the domination of noisy data. After NT training,  we train the model with only the samples having NT confidence over $\frac{1}{K}$, where $K$ denotes the number of classes. We refer to this process as Selective NT (SelNT), as illustrated in Figure \ref{fig:pt_nt} (c)   \cite{kim2019nlnl}. We also depict the distribution  of proposed RNT in Figure \ref{fig:pt_nt} (d), which  demonstrates the improvement of RNT in terms of noise filtering.
In terms of performance, as shown in Figure \ref{fig:pt_nt_performace}, the accuracy of PT on the Dev data increases in the early stage. However, the direct mapping of features to the noisy labels  eventually leads to the overfitting problem and thus gradually results in inaccurate performance on the clean Dev data.
\subsection{Noise Ranking} \label{sec:ranking}

We begin by extracting the shared features (i.e., evidential support) between the evidences (i.e., the labeled texts) and  the inference (i.e., the unlabeled texts). 
Then, we adopt a reasoning with uncertainty approach to measure the evidential support. The instance with higher evidential support is regarded as a less potential noisy instance.  An illustrative example is shown in Figure \ref{fig:DST}. Next, we describe the process in detail. 

\subsubsection{Feature Generation }

Recall that RNT begins by training on the labeled data using PT technique.  Consequently, we rely on the learned latent space of PT to generate various features with three properties, including automatically generated, discriminating, and high-coverage, as follows. 
\\
\textbf{Semantic distance}.
For each instance $x_i \in \{D_l,D_u\}$, we recompute its semantic relatedness to each label $y_i \in \mathcal{Y}$ based on the Angular Margin (AM) loss  \cite{wang2018cosface}. 
 
The AM loss adds a margin term to the Softmax loss based on the angle (i.e., cosine similarity) between an input sample's feature vector and the actual class's weight vector. Notably, the margin term encourages the network to learn feature representations that are well-separated and distinct for different classes. As a result, the angle between the feature vectors of an input sample and different classes becomes an essential factor in estimating the degree of noisiness. For clarity, we first describe the AM loss with respect to angles. Given a training example $(x_i; y_i)$, it can be formulated as:
	\begin{equation}
		\resizebox{\linewidth}{!}{$\mathcal{L}_{cos} (x_i,y_i, \phi) = -  \sum_{k=1}^{K} y_{ik} \log( \frac{e^{s(cos(\theta_{ik}-y_{ik}m))}}{\sum_{j=1}^{K}e^{s(cos(\theta_{ij}-y_{ij}m))}})$},
	\end{equation}
	where $\phi$ denotes the model parameters and $cos(.)$ stands for the cosine similarity, which can be read as the angular distance between feature vectors and the class weights. Given an unlabeled instance $x^u_j$, we recompute its AM loss to each class $y_i \in \mathcal{Y}$ as follows:

\begin{equation}\label{eq:am_feature}
	\mathcal{L}_{cos} (x^u_j,y_i, \phi) = - \frac{1}{N} \sum_{n=1}^{N} \sum_{k=1}^{K} y_{ik} \log(\theta_{jn}),
\end{equation}
where $N$ is the number of samples (e.g., 5) from $D_l$ labeled with $y_i$ (i.e., class on-target) and $\theta_{jn}$ denotes the cosine similarity between $x^u_j$ and $x^l_n$ (i.e., the deep representations of the PT classifier). The intuition behind this feature is that an unlabeled instance $x_j^u$, that receives close amount of support from different classes, is regarded as potential mislabeled. We denote this feature as $f$ and its value consists  of the corresponding class $y_i$ as well as the value of $\mathcal{L}_{cos}$. To enable valuable shared knowledge between instances, $\mathcal{L}_{cos}$ is approximated to one digit (e.g., $\mathcal{L}_{cos}(x^u_j, 1)=0.213\approx0.2$). Considering the illustrative example in Figure \ref{fig:DST}, $x_j^u$ and  $x_i^l$ approximately share the same   $\mathcal{L}_{cos}$ in response to class $0$, (i.e., $f_3(0,0.2)$).

\textbf{PT confidence}.
Instances with extreme confidence (i.e., close to 1) are generally considered to have a low risk of being mislabeled  \cite{hou2018r}. To incorporate the class distribution of PT into the evidential support measurement process, we introduce a new feature, denoted as $f'$, whose value consists of the predicted class and its corresponding probability. Considering the illustrative example in Figure \ref{fig:DST}, $x_j^u$  and $x_i^l$ share $f'_1$ (i.e., $f'_1(0,0.9)$), which can be read as both instances are related to the class 0 based on PT classifier with $0.9$ confidence.

\subsubsection{Evidential Support Measurement }

Now that we can capture shared knowledge between the labeled and unlabeled instances (i.e., the evidential support). {  We leverage Dempster Shafer Theory (DST)  \cite{yang2013evidential} to address evidential support measurement as reasoning with uncertainty. The goal is to estimate the degree of noisiness for an unlabeled instance by combining its evidence from multiple sources of uncertain information (i.e., PT and semantic features). To achieve this, DST applies Dempster's rule, which combines the mass functions of each source of evidence to form a joint mass function. } ot is noteworthy that DST has been widely used for various purposes of reasoning  \cite{Liu2018Classifier,wang2019gradual,ahmed-etal-2021-dnn}. The basic concepts of DST are:
\begin{itemize}
	\item \textbf{Proposition}. It refers to all possible states of a situation under consideration. Two propositions are defined: ``clean instance'', denoted by $C$, and ``unclean instance'', denoted by $U$. Let proposition be  $X = \{C, U\}$ and a power set of $X$ be $2^X = \{\emptyset,C, U, X\}$.
	
	\item \textbf{Belief function}. It associates each $E \in 2^X$ with a degree of belief (or mass), which satisfies $\sum_{E\in 2^X} {m(E)=1}$ and $m(\emptyset )=0$. 	Different belief functions for various evidences are defined (i.e., the generated features). 
\end{itemize}

\begin{figure}[t]
	\centering
	\includegraphics[scale=.47]{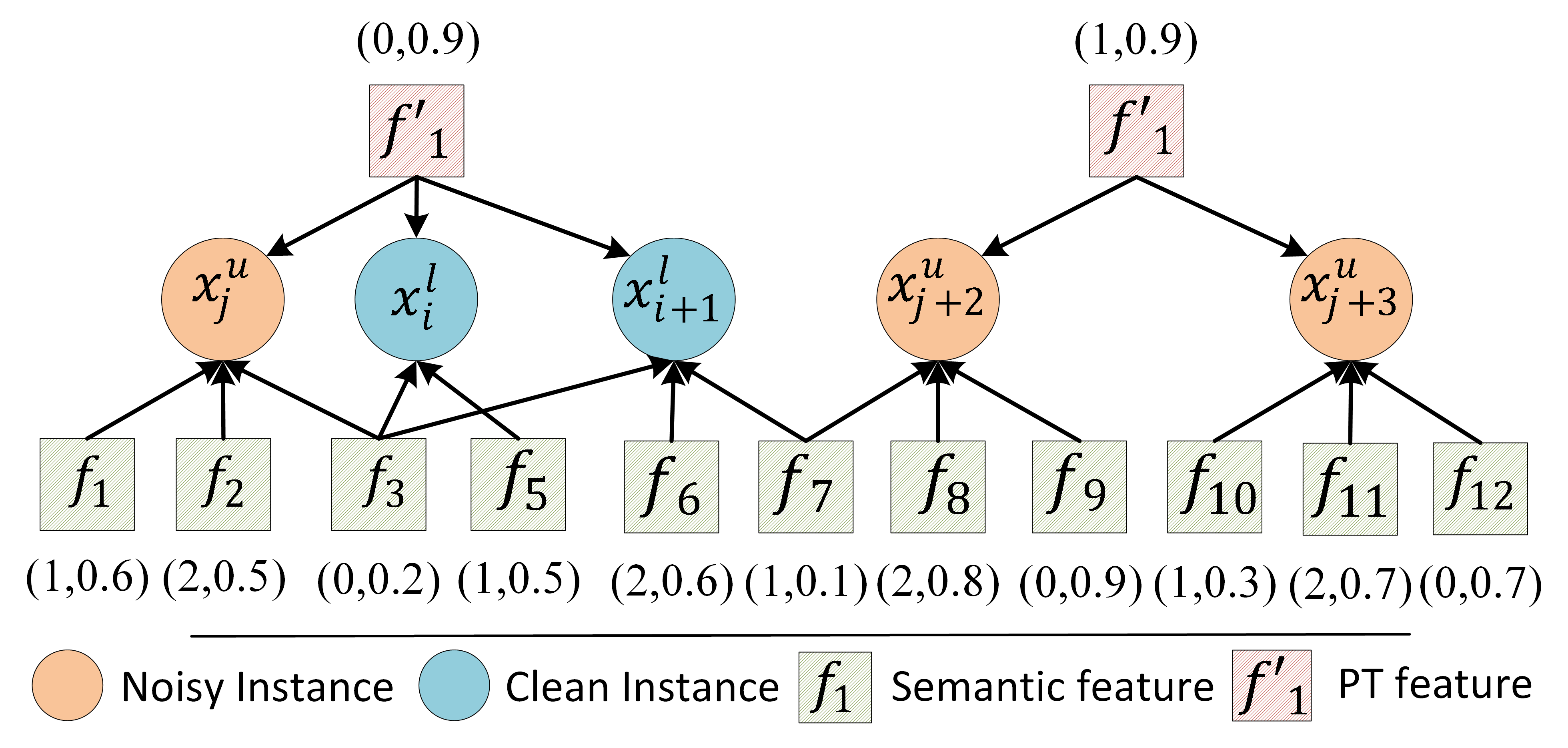}	
	\caption{An illustrative example of the evidential support. The instances $x_j^u$ and $x_i^l$ exhibit a similar $\mathcal{L}_{cos}$ value in response to class $0$ (i.e., $f_3(0,0.2)$). The approximate PT's confidence of 0.9, represented by $f'_1(0,0.9)$, further strengthens this similarity. Consequently, the instance $x_j^u$ is considered less noisy due to the higher degree of evidential support it receives.}
	\label{fig:DST}
\end{figure}

Given an unlabeled instance $x_j^u $ and its  semantic feature $f$, we estimate the evidential support that $x_j^u$ receives from labeled instances that share $f$ by the belief function:

\resizebox{7.5cm}{!}{
	\begin{minipage}{\linewidth}
		\begin{align*}
			m_{f}(E)= 
			\left \{
			\begin{array}{lr}
				(1 - d_f )  \max\{P(f ), 1 - P(f )\} & E =\{C\} \\
				(1 - d_f )  \min\{P(f ), 1 - P(f )\} &	E= \{U\} \\
				d_f & 	E= \{C,U\}
			\end{array}
			\right.\\
		\end{align*}
	\end{minipage}}
\vspace{-20pt}
\begin{equation} \label{eq:bf}	
\end{equation}
where $d_f$ denotes the degree of uncertainty of $f$, and $P(f)$ { is the division of the number of positive instances (i.e.,  the labeled instances with the same class of the feature on-target  $f$) by all labeled instances shared  $f$. Consider the illustrative example in Figure \ref{fig:DST}, $f_3(0,0.2)$ (i.e., semantically related to class 0 with approximated similarity of 0.2), suppose that the positive instances $x_i^l$ and $x_{i+1}^l$ are annotated with class $0$, then $P({f_3})=1.0$.}
Equation \ref{eq:bf} can be read as  the more extreme the value of $P(f)$ (i.e. close to 0 or 1) is, the more evidential support the element of ${C}$ should receive from the feature $f$. Similarly, we use Equation \ref{eq:bf} to estimate the evidential support $m_{f'}(E)$ that $x_i^u$ receives from  $f'$. Note that $d_f$ represents the impact that a given feature may have on the final degree of belief in terms of evidential support measurement. The lower the value, the greater the impact.  Note that both types of features are generated based on the latent space of the PT classifier that we believe in its semantic representation as it is trained on the labeled data. Therefore, we empirically set $d_f$ to a small unified value (i.e., $0.2$ in our experiments).

The overall evidential support of $E=\{C\}$ that  $x_j^u$ receives from its observations is  estimated by combining the estimated beliefs
as follows:

\begin{equation}\label{eq:estimate_es}
	m(E) = m_{f_1}(E) \oplus ... \oplus m_{f_n}(E) \oplus m_{f'}(E),
\end{equation}
where $m(E)$ represents the total amount of evidential support that $x_j^u$ receives, and the combination is computed from the two sets of belief functions, $m_{f_1}(E)$ and  $m_{f_2}(E)$  as follows:
\begin{equation} 
				\resizebox{\linewidth}{!}{$	m_{f_1}(E) \oplus m_{f_2}(E) =\frac{1}{1-U}  \sum_{E' \cap E'' =E}  m_{f_1}(E') m_{f_2}(E'')$}
			\end{equation}
			where $E'$ and $E''$ denote the power set $2^X$ elements and $U = \sum_{E' \cap E'' =\emptyset} m_{f_1}(E') m_{f_2}(E'')$  is a measure of the amount of conflict between  $E'$ and $E''$. 
			In words, given the element of $E=\{C\}$, we multiply the combinations of $E'$ and $E''$ such that $E'\cap E''={C}$ {and thus can be regarded as a measure for the amount of support from $\{C\}$. }
			For time complexity, each iteration takes $O(n \times n_f )$ time with $n$ instances and $n_f$ the number of the generated features. Thus, the time complexity can be represented by $O(n^2 \times n_f)$.
			
			\subsection{Training Procedure}

			Now that we can measure the evidential support, we then rank the instances of $D_u$ and select the less risky instances as the filtered set, denoted as  $D'_u =\{(x_j^u,y_j^u )\}^{j=N_f}_{i=1}$. Note that the value of $N_f$ is fine-tuned using the Dev set (please refer to Section \ref{sec:exp} for more details). Finally, we combine both sets $D_l$ and $D'_u$ together for the final NT training, as illustrated in Figure \ref{fig:framework}. 
			The training procedure can be explained by the following steps. We first generate pseudo-labels using PT technique Eq.\ref{eq:pl}. Then, we apply DST to filter the highly risky instances. Finally, we adopt NT technique Eq.\ref{eq:nt} to alleviate the noisy information during the training.  Furthermore, to improve the convergence after NT, we follow  \cite{kim2019nlnl} by training only with the instances whose confidence is over $\frac{1}{K}$, denoted as SelNT in Figure \ref{fig:pt_nt} (c).

\section{Experimental Setup}\label{sec:experiments}
\subsection{Dataset}
We validate the performance of the proposed RNT on various text classification benchmark datasets Table \ref{tab:dataset_sst}. Particular, we rely on AG News  \cite{zhang2015character}, Yahoo  \cite{chang2008importance}, Yelp  \cite{zhang2015character}, DBPedia  \cite{zhang2015character}, TREC  \cite{li2002learning}, SST  \cite{socher2013parsing}, CR  \cite{ding2008holistic}, MR  \cite{pang2005seeing}, TNEWS, and  OCNLI  \cite{xu2020clue}.
For AG News, Yelp and Yahoo datasets, we follow the comparative approaches by forming the unlabeled training set $D_u$, labeled training set $D_l$ and development set by randomly drawing from the corresponding original training datasets. For the other datasets, we split the training set into  10\% and 90\% for  $D_l$ and $D_u$, respectively. Note that we utilize the original test sets for prediction evaluation.

\subsubsection{Comparative Baselines}
For fairness, we only include the semi-supervised learning methods that were built upon the contextual embedding models (e.g., BERT):

\begin{itemize}
	\item \textbf{PLM } is a pre-trained language model directly fine-tuned on the labeled data. We compared to BERT  \cite{devlin2018bert,cui2021pre} and RoBERTa  \cite{liu2019roberta};
	\item \textbf{UDA}  \cite{xie2020unsupervised} is an SSTC method based on unsupervised data augmentation with back translation. We use German and English languages for back-translation of English and Chinese, respectively,  datasets;
	\item \textbf{UST}  \cite{mukherjee2020uncertainty} introduces to select samples by information gain and utilizes cross-entropy loss to perform self-training; 
	\item \textbf{S$^2$TC-BDD }  \cite{li2021semi} is an SSTC method that addresses the margin bias problem  by balancing the label angle variances.
\end{itemize}

\begin{table*}
	\centering
	\resizebox{16 cm}{!}{
			\begin{tabular}{lcccccccll}
				\hline
				\multirow{2}{*}{Dataset}	&\multirow{2}{*}{\#Class} 	&\multicolumn{2}{c}{Train}	&\multirow{2}{*}{\#Dev} 	&\multirow{2}{*}{\#Test} 		& \multirow{2}{*}{Length}&\multirow{2}{*}{Language}&\makecell[c]{\multirow{2}{*}{Task}}&\makecell[c]{\multirow{2}{*}{Metric}}	\\\cline{3-4}
				&			&\#Lab 	&\#Unlab			&		&			\\
				\hline
				AG News & 4		&10k	&20k	&8k		&7.6k 	&100&English&Topic 		&Macro-F1	\\
				Yelp 	& 5 	&10k 	&20k	&10k	&5k		&256&English&Sentiment 	&Macro-F1	\\
				Yahoo	& 10 	&10k 	&40k	&20k	&60k	&256&English&Topic 		&Macro-F1	\\
				DBPedia	& 14 	&10k 	&20k	&10k	&70k 	&160&English&Topic 		&Macro-F1	\\
				TREC	&6		&5.4k	&NA		&1.1k	&500 	&30&English&Question  	&Macro-F1	\\
				SST		&\{2,5\}&6.9k 	&NA 	&871 	&1.8k	&50&English&Sentiment 	&Macro-F1	\\
				CR	 	&2		&3k		&NA		&378	&372	&50&English&Sentiment 	&Macro-F1	\\
				MR		&2		&6.9k	&NA		&1.7k	&2k		&50&English&Sentiment 	&Macro-F1	\\			
				TNEWS 	&15		&53.3k	&NA		& 10k	& 10k	&128&Chinese& Topic  		& Accuracy\\
				OCNLI 	&3		&50k 	&NA		&3k		& 3k	&128&Chinese& NLI & Accuracy\\
				\hline
			\end{tabular}
		}
		\caption{The statistics of benchmark datasets, whereas \#Lab and \#Unlab denote the number of labeled and unlabeled texts, respectively. Note that for datasets with NA, we split  \#Lab into 10\% and 90\% for \#Lab and \#Unlab, respectively.   }
		\label{tab:dataset_sst}
	\end{table*}

\subsection{Experimental Settings} \label{sec:exp}
\begin{figure}[t]
	\centering
	\includegraphics[scale=.46]{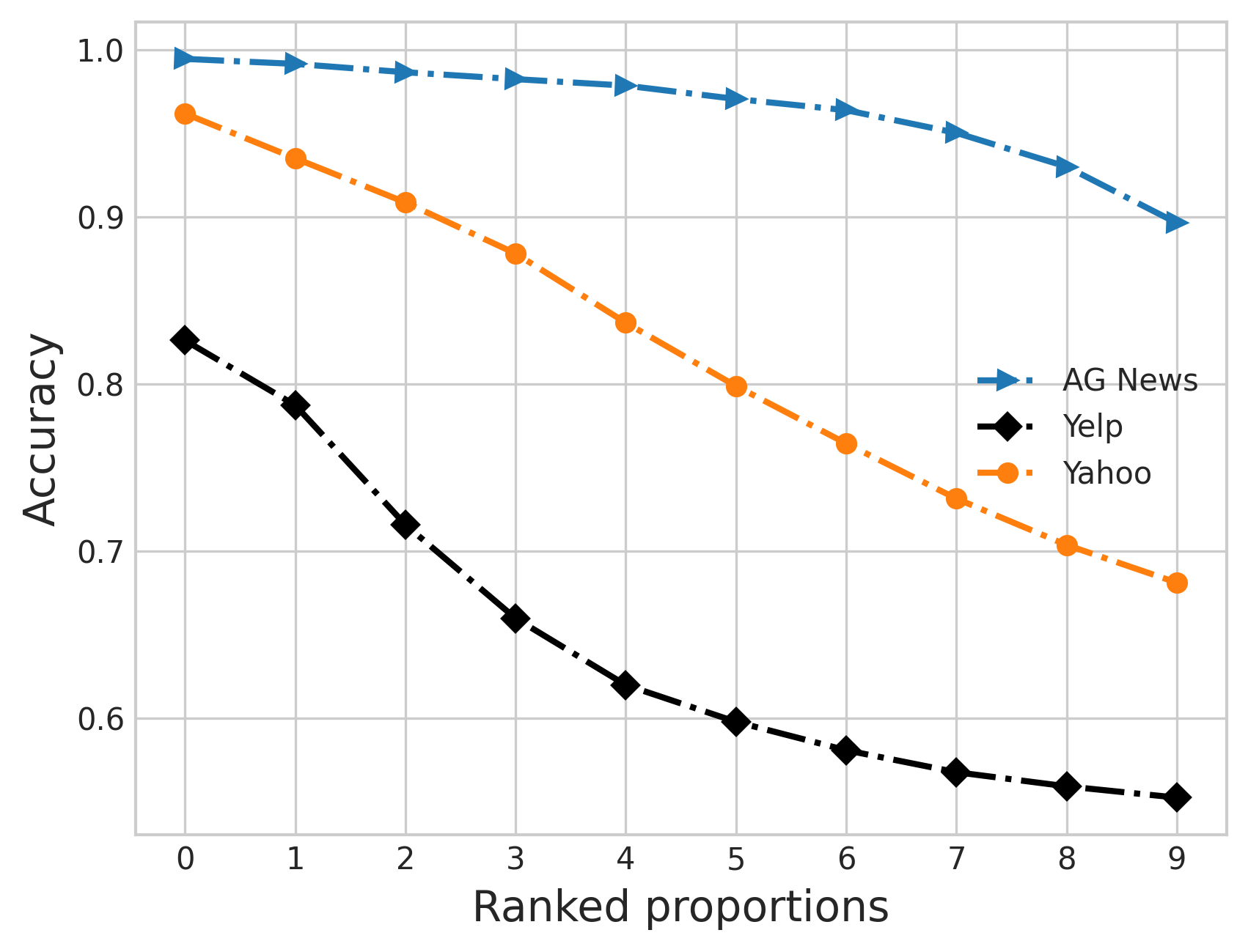}	
	\caption{Ranking evaluation on Dev sets with $N_l=1k$. The ranked Dev set is first split into 10 proportions equally-likely. {Then, each proportion is inferred (i.e., calculate its accuracy) independently}. The accuracy gradually drops as the noisy texts increase. In our experiment, we choose the proportions whose instances meet $\lambda$ Section \ref{sec:exp} for further NT training.    }
	\label{fig:dst_acc}
\end{figure}

\begin{itemize}
	
 \item \textbf{Hyper-parameters.} We use 12 heads and layers and keep the dropout probability to 0.1 with 30 epochs, learning rate of $2e^{-5}$  and 32 batch size. To guarantee the re-productivity without manual effort, we rely on the Dev set to automatically set the value of $N_f$ (i.e., the number of instances in $D'_u$). First, the ranked Dev set is split into small proportions (i.e., max is 10). Then, $m$ is set to proportions that meet the condition $\lambda =max(p)-st(p)$, where $p$ is a vector that represents the accuracy of RNT on each proportion and $st$ denotes the standard deviation. For example, $\theta=0.2$ means that  $D'_u$ consists of the first 20\% of the ranked $D_u$ as shown in Figure \ref{fig:dst_acc}. We set the number of negative samples to $K-1$, where $K$ is the number of classes in the labeled training set.
\item \textbf{Metrics.}  We use the accuracy metric on Clue datasets, including TNEWS and  OCNLI, while Macro-F1 scores for all other datasets. 
\end{itemize}

\section{Evaluation and Results}
We describe the evaluation tasks and report the experimental results in this section. The evaluation criteria are:
(I) Is RNT able to rank the instances of being mislabeled?; 
(II) Can the filtered data enhance the performance of the clean test data?

\begin{table*}[t]
	\centering
	\resizebox{16 cm}{!}{
			\begin{tabular}{clccccccccccccc}
				\hline
				\multirow{2}{*}{PLM}	&\multirow{2}{*}{Model}&\multicolumn{3}{c}{AG News}&&\multicolumn{3}{c}{Yelp}&&\multicolumn{3}{c}{Yahoo} &&DBPedia\\\cline{3-5}  \cline{7-9}  \cline{11-13} \cline{15-15} 
				&&30&1k&10k&&30&1k&10k&&30&1k&10k&&30\\
				\hline
				
				\multirow{5}{*}{\makecell[c]{BERT-\\Base}}
				&Fine-tuning  	&84.1$\pm$0.9			&87.8$\pm$0.3			&90.5$\pm$0.2			&&42.2$\pm$1.7			&53.2$\pm$0.8			&58.6$\pm$0.5				&&63.2$\pm$0.5			&67.1$\pm$0.3			&70.8$\pm$0.2			&&97.1$\pm$0.9\\
				&UDA	 		&85.7$\pm$0.3			&88.3					&90.6					&&44.6$\pm$1.2			&55.0					&57.6					&&66.4$\pm$0.5			&66.6					&70.4					&&98.5$\pm$0.6\\
				&UST			&\textbf{87.2}$\pm$0.6	&88.6					&90.8					&&44.8$\pm$1.1			&54.2					&57.7					&&\textbf{66.5}$\pm$0.3	&67.5					&71.1					&&98.4$\pm$0.6\\
				&S$^2$TC-BDD 	&86.9$\pm$0.7			&88.9					&90.7					&&\textbf{45.9}$\pm$1.4	&55.0					&58.6					&&66.2$\pm$0.6			&68.0					&70.9					&&\textbf{98.8}$\pm$0.7\\
				
				&RNT (Ours)		&86.7$\pm$0.3			&\textbf{89.4}$\pm$0.1	&\textbf{91.9}$\pm$0.1	&&44.9$\pm$1.2			&\textbf{56.6}$\pm$0.6	&\textbf{60.2}$\pm$0.1	&&66.2$\pm$0.3			&\textbf{69.1}$\pm$0.2	&\textbf{72.7}$\pm$0.1	&&98.2$\pm$0.4\\
				\hline	
				\multirow{2}{*}{\makecell[c]{RoBERTa-\\Base}}			
				&Fine-tuning  	&84.9$\pm$0.7			&88.5$\pm$0.3			&91.0$\pm$0.2			&&53.7$\pm$1.6			&57.8$\pm$0.7			&62.5$\pm$0.4			&&66.6$\pm$0.4			&68.3$\pm$0.5			&72.3$\pm$0.2			&&98.1$\pm$0.5	\\								
				&RNT (Ours)		&\textbf{86.9}$\pm$0.4	&\textbf{89.6}$\pm$0.1	&\textbf{92.2}$\pm$0.2	&&\textbf{53.9}$\pm$1.4	&\textbf{60.0}$\pm$0.5	&63.8$\pm$0.1			&&\textbf{67.2}$\pm$0.4		&\textbf{69.6}$\pm$0.2			&\textbf{73.7}$\pm$0.1			&&98.4$\pm$0.2\\
				\hline	
				\multirow{2}{*}{\makecell[c]{RoBERTa-\\Large}}	
				&Fine-tuning	&86.5$\pm$0.4			&89.1$\pm$0.2			&91.8$\pm$0.2			&&56.2$\pm$1.3			&62.3$\pm$0.6			&66.0$\pm$0.4			&&67.8$\pm$0.3			&70.3$\pm$0.3			&73.7$\pm$0.1	&&98.3$\pm$0.3\\						
				
				&RNT (Ours) 	&\textbf{87.8}$\pm$0.3	&\textbf{89.8}$\pm$0.1	&\textbf{92.6}$\pm$0.1	&&\textbf{58.3}$\pm$0.9	&\textbf{63.1}$\pm$0.4	&\textbf{66.8}$\pm$0.2	&&\textbf{68.9}$\pm$0.2	&\textbf{71.2}$\pm$0.2			&\textbf{74.3}$\pm$0.1			&&98.8$\pm$0.2\\
				\hline									
				
		\end{tabular}}
		\caption{Comparative results with the state-of-the-art alternatives on 30 examples per label and $N_l\in\{1k,10k\}$. The results of  $N_l\in\{1k,10k\}$ are retrieved from S$^2$TC-BDD  \cite{li2021semi}, while the others are our implementations. The scores consists of the average of three runs, and the best scores are in bold. }
		\label{tab:res_sstc}
	\end{table*}

\begin{table*}
	\centering

		\resizebox{16 cm}{!}{
		\begin{tabular}{clcccccccccc}
			\hline		
			\multirow{2}{*}{PLM}&\multirow{2}{*}{Model}&\multicolumn{2}{c}{TREC}&\multicolumn{2}{c}{SST-2}&\multicolumn{2}{c}{SST-5}&\multicolumn{2}{c}{CR}&\multicolumn{2}{c}{MR}\\ \cline{3-12}
			&&30&10\%		&30&10\%		&30&10\%		&30&10\%		&30&10\%	\\
			
			\hline
\multirow{5}{*}{\makecell[c]{BERT-\\ Base}}
& Fine-tuning 	& 78.7$\pm$1.6 			& 87.1$\pm$1.0 			& 76.9$\pm$1.6 			& 85.2$\pm$0.8 			& 33.2$\pm$1.4 			& 39.0$\pm$1.1 			& 74.7$\pm$1.2 			& 85.8$\pm$0.9 			& 66.6$\pm$1.4 			& 80.7$\pm$0.7\\						
& UDA			& 83.5$\pm$1.1 			& 91.2$\pm$0.7 			& 79.9$\pm$1.3 			& 85.6$\pm$0.3			& 33.6$\pm$1.1	 		& 40.6$\pm$0.8			& 81.0$\pm$0.7 			& 87.7$\pm$0.6 			& \textbf{72.9}$\pm$0.9 & 81.0$\pm$0.1\\
& UST 			& 83.3$\pm$1.2			& \textbf{92.1}$\pm$0.8	& 78.7$\pm$1.0 			& 85.6$\pm$0.4 			& 33.9$\pm$1.1	 		& 40.8$\pm$0.7 			& \textbf{82.7}$\pm$0.8 & 87.8$\pm$0.3 			& 71.1$\pm$1.0			& 81.0$\pm$0.3\\
& S$^2$TC-BDD	& 81.2$\pm$1.3			& 91.2$\pm$0.9 			& 81.1$\pm$1.2 			& 85.7$\pm$0.5 			& 34.6$\pm$1.3 			& 39.6$\pm$0.5 			& 82.3$\pm$0.9 			& 87.6$\pm$0.7 			& 72.1$\pm$0.9 			& 80.0$\pm$0.6\\
& RNT (Ours)	& \textbf{85.2}$\pm$1.1 & 91.4$\pm$0.7  		& \textbf{83.8}$\pm$1.3 & \textbf{87.6}$\pm$0.4 & \textbf{35.9}$\pm$1.2 & \textbf{42.3}$\pm$0.9 & 82.6$\pm$0.9 			& \textbf{89.3}$\pm$0.4 & 71.5$\pm$1.0 & \textbf{82.4}$\pm$0.3\\						
\hline
\multirow{2}{*}{\makecell[c]{RoBERTa-\\Base}}

& Fine-tuning  	& 84.2$\pm$1.3			 & 92.1$\pm$0.7 		& 85.0$\pm$0.9 			& 89.5$\pm$0.4 			& 39.3$\pm$1.0 			& 47.6$\pm$0.7 			& 86.5$\pm$1.2 			& 91.1$\pm$0.7 			& 71.2$\pm$1.4 			& 84.9$\pm$0.5\\					
& RNT (Ours)	& \textbf{86.7}$\pm$0.8  & \textbf{93.2}$\pm$0.4& \textbf{87.7}$\pm$0.7 & \textbf{90.7}$\pm$0.4 & \textbf{40.5}$\pm$0.6 & \textbf{49.6}$\pm$0.4 & \textbf{88.9}$\pm$0.6 & \textbf{92.5}$\pm$0.2 & \textbf{75.8}$\pm$0.7 & \textbf{86.4}$\pm$0.2\\

\hline
\multirow{2}{*}{\makecell[c]{RoBERTa-\\Large}}
& Fine-tuning  	& 88.9$\pm$1.1 			& 92.5$\pm$0.6 			& 87.7$\pm$1.0			& 92.3$\pm$0.7 			& 40.5$\pm$0.8 			& 51.0$\pm$0.6 			& 89.7$\pm$0.9 			& 91.8$\pm$0.8 			& 82.4$\pm$1.2 			& 88.3$\pm$0.6\\
& RNT (Ours)	& \textbf{89.6}$\pm$0.6 & \textbf{94.0}$\pm$0.4 & \textbf{89.6}$\pm$0.6 & \textbf{93.2}$\pm$0.5 & \textbf{42.8}$\pm$0.5 & \textbf{52.4}$\pm$0.3 & \textbf{92.3}$\pm$0.6 & \textbf{92.6}$\pm$0.3 & \textbf{85.9}$\pm$0.6 & \textbf{88.4}$\pm$0.3\\
\hline

		\end{tabular}
	}
	\caption{Comparative results with the state-of-the-art alternatives with 30 samples per label and $N_l= 10\%$ of the labeled texts. Note that all results are the average of three runs with different seeds. }
	\label{tab:res_sstc_2}
\end{table*}

\begin{table}
	\centering
		\begin{tabular}{lcc}
			\hline		
			\multirow{2}{*}{Model}&TNEWS&OCNLI 	\\ \cline{2-3}
			&10\%&10\%\\
			
			\hline
			Fine-tuning 	&53.9&62.6\\
			UDA 	 		&52.3&63.8\\
			UST			  	&54.3&63.7\\
			S$^2$TC-BDD   	&53.4&64.5\\			
			RNT (Ours)		&\textbf{54.6}$\pm$0.4&\textbf{65.2}$\pm$0.3\\
			\hline
			
	\end{tabular}
	\caption{ Comparative results on Chinese datasets based on initial weights from RoBERTa-Large  \cite{cui-etal-2020-revisiting}. The best scores are in bold. }
	\label{tab:res_Chinese}
\end{table}

\subsection{Results}
We use the Dev set to select the best model and average three runs with different seeds. The experimental results are reported in Tables \ref{tab:res_sstc}, \ref{tab:res_sstc_2} and \ref{tab:res_Chinese} from which we have made the following observations.  
\begin{itemize}
\item  \textbf{Compared to the baselines}, RNT gives the best results compared to its alternatives in most cases and achieves competitive performance in others. We also observe that SSTC-based approaches comfortably outperform the PLMs fine-tuning when training with scarce labeled data (e.g., $N_l =30$); however, the same performance is expected when $N_l$ is increased (e.g., $N_l\in\{1k,10k\}$), but it was not supported by the experiments. Furthermore, experimental results demonstrate that  RNT is not sensitive to the number of classes compared to SSTC-based alternatives. For instance, UDA \cite{xie2020unsupervised} can perform better on the binary datasets as shown in Table \ref{tab:res_sstc_2}.
\item \textbf{  \textbf{Compared to the PLM}} fine-tuned on the labeled data, RNT comfortably overcomes PLMs by considerable margins. For example, the Macro-F1 scores of RNT with $N_l = 30$ are even about 2.6\%, 2.7\% and 3.0\% on AG News, Yelp and Yahoo datasets, respectively. Moreover, we also observe that RNT can perform better than PLM fine-tuned on sufficient labeled data (e.g., $N_l=10k$). 

\end{itemize}

\subsection{Mislabeling Filtering Evaluation}

To evaluate the ability of RNT  in mislabeling filtering, we conduct experiments on the Dev sets of AG News, Yelp, and Yahoo datasets as follows. We first associate the instances with the corresponding pseudo-labels (i.e., inferring using the PT classifier). Then, we require RNT to rank them based on their evidential support received from the clean training set (i.e., $N_l=1k$). Since we have access to the true labels of the Dev set, we can evaluate the performance of the filtering process. Specifically, we divide the ranked Dev set into ten equally-likely proportions (note that we keep the same order of ranking) and calculate the accuracy of each proportion separately (i.e., comparing the pseudo-labels with the true labels).
  The proportions, as shown in Figure \ref{fig:dst_acc}, are significantly correlated with the extent of mislabeling.
 In other words, the accuracy score gradually drops as the mislabeled instances increase and vice-versa. Note that we report the accuracy due to the imbalance labels in the proportions. Moreover, we report the performance of both the full Dev set and the filtered set in Table \ref{tab:Ranking_evaluation}.    
	 
\subsection{The Impact of Noise Filtering}	
	To assess the impact of noise filtering on the overall performance of RNT, we remove DST and conduct experiments on the AG News, Yelp, and Yahoo datasets. 
	The experimental results presented in Table \ref{tab:ablation} show that removing noise ranking from RNT causes a performance drop of 1.3, 1.6, and 1.2 on the AG News, Yelp, and Yahoo datasets, respectively. This demonstrates the efficacy of a well-designed noisy ranking in improving text classification performance
	Furthermore, we observe that even without noise filtering, RNT outperforms PLM fine-tuning and achieves competitive results compared to other alternatives. This supports the adoption of NT for noisy data.
	\begin{table*}[h]
	\centering
	\resizebox{16 cm}{!}{
			\begin{tabular}{lccccccccccc}
				\hline
				\multirow{2}{*}{Model}&\multicolumn{3}{c}{AG News}&&\multicolumn{3}{c}{Yelp}&&\multicolumn{3}{c}{Yahoo}\\\cline{2-4}  \cline{6-8}  \cline{10-12}
				&30&1k&10k&&30&1k&10k&&30&1k&10k\\
				\hline
				RNT Pure 		&82.9$\pm$0.8			&88.1$\pm$0.2			&91.3$\pm$0.2			&&42.6$\pm$1.7			&55.0$\pm$0.8			&59.8$\pm$0.3				&&65.1$\pm$0.7				&67.9$\pm$0.2			&71.9$\pm$0.1\\
				RNT PT-conf		&83.7$\pm$1.3			&\textbf{89.7}$\pm$0.2	&91.7$\pm$0.1			&&42.2$\pm$1.6			&54.6$\pm$0.6			&60.1$\pm$0.2				&&65.4$\pm$1.0				&68.1$\pm$0.4			&72.3$\pm$0.1\\
				
				RNT (Ours)		&\textbf{86.7}$\pm$0.3	&{89.4}$\pm$0.1			&\textbf{91.9}$\pm$0.1	&&\textbf{44.9}$\pm$1.2	&\textbf{56.6}$\pm$0.6	&\textbf{60.2}$\pm$0.1	&&\textbf{66.2}$\pm$0.3		&\textbf{69.1}$\pm$0.2	&\textbf{72.7}$\pm$0.1\\
				\hline				
		\end{tabular}}
		\caption{The effect of DST on the performance of RNT. All variants are jointly trained on $D_l$ and $D_u$ using PT and NT.  RNT Pure is trained on all instances in $D_u$ without any filtering mechanism, while RNT PT-conf uses the PT-based confidence to filter the instances in $D_u$ that does not meet a predefined threshold (i.e., 0.9 in our experiments).  }
		\label{tab:DST_pt_evaluation}
	\end{table*}
\begin{table}[t]
	\centering
	\resizebox{7.8cm}{!}{
		\begin{tabular}{llcccccc}
			\hline
			\multirow{2}{*}{Dataset}&\multirow{2}{*}{$N_l$} &\multicolumn{2}{c}{Full Dev} && \multicolumn{3}{c}{Filtering}	\\\cline{3-4}\cline{6-8}
			&&Acc&F1&& Prop&Acc&F1\\
			\hline
			\multirow{2}{*}{\makecell[l]{AG \\News}}&
			1k&88.1&88.1&&70\%&95.8&95.2 \\
			&10k&91.9&91.9&&70\%&98.2&97.9\\
			\hline
			\multirow{2}{*}{Yelp}&
			1k&53.8&53.2&&30\%&68.8&64.2 \\
			&10k&60.5&60.2&&30\%&75.9&72.5 \\
			\hline
			\multirow{2}{*}{Yahoo}&
			1k&67.1&67.0&&30\%&89.6&72.6\\
			&10k&72.0&71.2&&40\%&91.0&83.6 \\
			\hline
	\end{tabular}}
	\caption{ Filtering evaluation on Dev sets. Prop, Acc and F1 denote proportion (i.e., the ratio of filtered texts), accuracy and Macro-F1, respectively. }
	\label{tab:Ranking_evaluation}
\end{table}

\subsection{The Effect of DST}
	
	To validate the contribution of DST on the final performance in terms of mislabeled instances filtering, we implement two variants, namely  RNT Pure and RNT PT-conf, as follows. The RNT Pure is trained on $D_l$ and $D_u$ as a whole without any filtering mechanism, while RNT PT-conf uses the PT confidence to filter the instances in $D_u$ that does not meet the predefined threshold (i.e., 0.9 in our experiments). In other words, instead of DST, we rely on the PT confidence to discard the instances close to the boundary. Empirically, we conduct experiments on AG NEWS, Yelp, and Yahoo datasets with various $N_l =\{30, 1k,10k\}$. The comparative results are shown in Table \ref{tab:DST_pt_evaluation} from which we made the following observations. Overall, RNT can mostly give the best performance, and the improvements are significant, especially with less limited data (e.g., $N_l=30$). RNT Pure performs worse due to the absence of a filtering mechanism. RNT PT-conf can achieve competitive performance with sufficient labeled data (e.g., $N_l=10k$) even in terms of uncertainty. However, it gradually drops with the decrease of labeled data. Intuitively, these results are expected as the performance of the PT classifier heavily relies on the amount of labeled data. In brief, the ablation study empirically supports the contribution of DST to the performance of RNT.

	\begin{table}
		\centering
			
			\begin{tabular}{lccc}
				\hline

				&AG News&Yelp&Yahoo\\
				\hline
				BERT (fine-tune)&87.8&53.2&67.1\\
				S$^2$TC-BDD&88.9&55.0&68.0\\
				\hline
				RNT w/o ranking &88.1&55.0&67.9\\
				RNT &89.4&56.6&69.1\\
				\hline
			\end{tabular}
		\caption{{  The impact of noise filtering to the overall performance. Note that the number of labeled data is set to $N_l=1k$. Removing noise ranking from RNT leads to a noticeable performance drop; however, it still performs better than BERT-Based fine-tuned on labeled data and achieves competitive scores comparable to S$^2$TC-BDD.} }
		\label{tab:ablation}
	\end{table}
	
\begin{table}[t]
	\centering
	\begin{tabular}{lcccc}
		\hline
		\multirow{2}{*}{Dataset} &\multicolumn{2}{c}{Full Dev} &&\multirow{2}{*}{\makecell[c]{Denoising\\ accuracy}}	\\\cline{2-3} 
		&clean&noise&& \\

		\hline
		AG News	&91.92&84.74&&87.53 \\
		Yelp	&60.49&57.14&&74.12 \\
		Yahoo	&71.96&68.46&&83.25 \\
		
		\hline
	\end{tabular}
	\caption{ { Denoising evaluation with $N_l=10k$ and 30\% of noise instances. Full Dev denotes the performance on the clean and noisy Dev sets. Denoising indicates the ability of RNT to identify the clean instances.   } }
	
	\label{tab:DST_evaluation}
\end{table}

	\subsection{Denoising Evaluation}

	Recall that the ultimate goal of DST is to estimate the score of unlabeled instances being mislabeled by the PT classifier. To evaluate the ability of DST to denoising, we adopt a perturbation strategy that has been used widely in the literature  \cite{belinkov2018synthetic,sun-jiang-2019-contextual}. We randomly pick 30\% of the Dev data as the noisy instances. For each instance, we randomly select 30\% of the words to be perturbed as follows. Specifically, we apply four kinds of noise: 
	(1) swap two letters  per word;
	(2) delete a letter randomly in the middle of the word;
	(3) replace a random letter with another in a word;
	(4) insert a random letter in the middle of the word.

	The evaluation results of denoising are reported in Table \ref{tab:DST_evaluation} from which we made the following observations. (1)  A considerable margin between the performance of the PT classifier on the clean and noise data, demonstrating the impact of the generated noise. (2)  Despite the well-recognized challenge of denoising in NLP, our proposed solution can mostly identify clean instances. (3) Even though the performance can be deemed considerable, noisy information may still exist in the filtered data; therefore, we use NT for further training.

\section{Conclusion and Future Work}\label{sec:conclusion}

In this paper, we proposed a self-training semi-supervised framework, namely RNT, to address text classification problem in learning with noisy labels manner. RNT first discards the high risky mislabeled texts based on  reasoning with uncertainty theory. Then, it uses the negative training technique to reduce the noisy information during training.
Our extensive experiments have shown that RNT mostly outperformed SSTC-based alternatives. Despite the robustness of negative training, clean samples that have identical distributions with test data are subjected to complementary labels. Consequently, both clean and potentially noisy samples contribute equally to the final performance.  A combination of both positive and negative training strategies in a unified framework can remedy the abundance of noisy samples; however, it needs further investigation.

\section*{Acknowledgements}
We extend our gratitude to the TACL action editor and the anonymous reviewers for their valuable feedback and insightful suggestions, which have significantly contributed to the improvement of our work.

\bibliographystyle{acl_natbib}
\bibliography{references}
\end{document}